\documentclass{article}
\usepackage{ijcai19}

\usepackage{times}
\usepackage{soul}
\usepackage{url}
\usepackage[hidelinks]{hyperref}
\usepackage[utf8]{inputenc}
\usepackage[small]{caption}
\usepackage{graphicx}
\usepackage{amsmath}
\usepackage{booktabs}
\usepackage[linesnumbered,ruled,vlined]{algorithm2e}
\usepackage{fancyvrb}
\usepackage{braket}
\title{\textbf{StepGAN} \\ Notes on Extending with Classes}
\usepackage{mathtools}
\usepackage{microtype}
\usepackage{nag}
\date{}

\newcommand{\qthd}{Q_\mathcal{D}(y_{1:t-1},  y_t)}
\newcommand{\qthc}{Q_\mathcal{C}(y_{1:t-1}, y_t, c)}
\newcommand{\vthd}{V_\mathcal{D}(y_{1:t-1})}
\newcommand{\vthc}{V_\mathcal{C}(y_{1:t-1,} c)}

\usepackage{amsmath,amssymb}

\usepackage{commath}
\title{GANs for Semi-Supervised Opinion Spam Detection\footnote{This paper has been accepted at IJCAI 2019.}}

\author{
Gray Stanton$^1$ \and
Athirai A. Irissappane $^2$\\
\affiliations
$^1$Colorado State University, gray.stanton@colostate.edu\\
$^2$University of Washington, athirai@uw.edu\\
}

\begin{document}

\maketitle

\begin{abstract}
 Online reviews have become a vital source of information in purchasing a service (product). Opinion spammers manipulate reviews, affecting the overall perception of the service. A key challenge in detecting opinion spam is obtaining ground truth. Though there exists a large set of reviews online, only a few of them have been labeled spam or non-spam. In this paper, we propose spamGAN, a generative adversarial network which relies on limited set of labeled data as well as unlabeled data for opinion spam detection. spamGAN improves the state-of-the-art GAN based techniques for text classification. Experiments on TripAdvisor dataset show that spamGAN outperforms existing spam detection techniques when limited labeled data is used. Apart from detecting spam reviews, spamGAN can also generate reviews with reasonable perplexity.
  
\end{abstract}

\section{Introduction}

Opinion spam is a widespread problem in e-commerce, social media, travel sites, movie review sites, etc.~\cite{Jindal2010FindingUR}. Statistics show that more than $90\%$ of the consumers read reviews before making a purchase~\cite{crowdstatistics}. The likelihood of purchase is also reported to increase when there are more reviews. Opinion spammers try to exploit such financial gains by providing spam reviews which influence readers and thereby affect sales. We consider the problem of identifying spam reviews as a classification problem, i.e., given a review, it needs to be classified either as spam or non-spam.

One of the main challenges in identifying spam reviews is the lack of labeled data, i.e., spam and non-spam labels~\cite{Rayana2015CollectiveOS}. While there exists a corpus of online reviews only few of them are labeled. This is mainly because manual labeling is often time consuming, costly and subjective~\cite{li2018generative}. Research shows that unlabeled data, when used in conjunction with small amounts of labeled data can produce considerable improvement in learning accuracy~\cite{Ott2011FindingDO}. There is very limited research on using semi-supervised learning techniques for opinion spam detection~\cite{crawford2015survey}. The existing semi-supervised learning approaches~\cite{li2011learning,hernandez2013using,li2014spotting} for identifying opinion spam use pre-defined set of features for training their classifier. In this paper, we will use deep neural networks which automatically discovers features needed for classification~\cite{lecun2015deep}.

Deep generative models have shown promising results for semi-supervised learning~\cite{kumar2017semi}. Specifically, Generative Adversarial Networks (GANs)~\cite{goodfellow2014generative} which have the ability to generate samples very close to real data, have achieved state-of-the art results. However, most research on GANs are for images (continuous values) and not text data (discrete values)~\cite{fedus2018maskgan}. 

GANs operate by training two neural networks which play a min-max game: discriminator D tries to discriminate real training samples from fake ones and generator G tries to generate fake training samples to fool the discriminator. The main drawback with GANs is that: 1) when the data is discrete, the gradient from the discriminator may not be useful for improving the generator. This is because, the slight change in weights brought forth by the gradients may not correspond to a suitable discrete mapping in the dictionary~\cite{huszar2015not}; 2) the discrimination is based on the entire sentence not parts of it, giving rise to the sparse rewards problem~\cite{yu2017seqgan}. 

Existing works on GANs for text data generation are limited by the length of the sentence that can be generated, e.g., MaskGAN~\cite{fedus2018maskgan} considers $40$ words per sentence. These approaches may not be suitable for processing most online reviews, which are relatively lengthy. For example, the TripAdvisor review dataset used in our experiments has sentences with median length $132$. Further, GANs have also not been fully investigated for text classification tasks.

In this paper, we propose spamGAN, a semi-supervised GAN based approach for classifying opinion spam. spamGAN uses both labeled instances and unlabeled data to correctly learn the input distribution, resulting in better prediction accuracy for comparatively longer reviews. spamGAN consists of $3$ different components: generator, discriminator, classifier which work together to not only classify spam reviews but also generate samples close to the train set. We conduct experiments on TripAdvisor dataset and show that spamGAN outperforms existing works when using limited labeled data. 

Following are the main contributions of this paper: 1) we propose spamGAN: a semi-supervised GAN based model to detect opinion spam. To the best of our knowledge, we are the first to explore the potential of GANs for spam detection; 2) the proposed GAN model improves the state-of-the-art GAN based models for semi-supervised text classification; 3) most existing research on opinion spam (other than deep learning methods) manually identify heuristics/features for classifying spamming behavior, however in our GAN based approach, the features are learned by the neural network; 4) experiments show that spamGAN outperforms state-of-the art methods in classifying spam when limited labeled data is used; 5) spamGAN can also generate spam/non-spam reviews very similar to the training set which can be used for synthetic data generation in cases with limited ground truth.

\section{Related Work}

Most existing opinion spam detection techniques are supervised methods based on pre-defined features. \cite{Jindal2008OpinionSA} used logistic regression with product, review and reviewer-centric features. \cite{Ott2011FindingDO} used n-gram features to train a Naive Bayes and SVM classifier. \cite{Feng2012SyntacticSF,mukherjee2013yelp,li2015analyzing} used part-of-speech tags and context free grammar parse trees, behavioral features, spatio-temproal features, respectively. \cite{wang2011review,akoglu2013opinion} used graph based algorithms. 

Neural network methods for spam detection consider the reviews as input wihtout specific feature extraction. GRNN~\cite{ren2017neural} used a gated recurrent neural network to study the contexual information of review sentences. DRI-RCNN~\cite{zhang2018dri} used a recurrent network for learning the contextual information of the words in the reviews. DRI-RCNN extends RCNN~\cite{lai2015recurrent} by learning embedding vectors with respect to both spam and non-spam labels for the words in the reviews. Since RCNN and DRI-RCNN use neural networks for spam classification, we will use these supervised methods for comparison in our experiments.

Few semi-supervised methods for opinion spam detection exist. \cite{li2011learning} used co-training with Naive-Bayes classifier on reviewer, product and review features. \cite{hernandez2013using,li2014spotting} used only positively labeled samples along with unlabeled data. \cite{Rayana2015CollectiveOS} used review features, timestamp, ratings as well as pairwise markov random field network of reviewers and product to build a supervised algorithm along with semi-supervised extensions. Other un-supervised methods for spam detection~\cite{xu2015unified} exists, but, they are out of the scope of this work.

The ongoing research on GANs for text classification aim to address the drawbacks of GANs in generating sentences with respect to the gradients and the sparse rewards problem. 
%(the computed loss is based on the entire sentence and not partially generated sequence, making it difficult to judge how good a partially generated sentence is when compared to the full sentence). 
SeqGAN~\cite{yu2017seqgan} addresses them by considering sequence generation as a reinforcement learning problem. Monte Carlo Tree Search (MCTS) is used to overcome the issue of sparse rewards, however it is computationally intractable. StepGAN~\cite{tuan2018improving} and MaskGAN~\cite{fedus2018maskgan} use the actor-critic~\cite{konda2000actor} method to learn the rewards, however MaskGAN is limited by length of the sequence. Further, all of them focus on sentence generation. CSGAN~\cite{li2018generative} deals with sentence classification, but it uses MCTS and character-level embeddings. spamGAN differs from CSGAN in using the actor-critic reinforcement learning method for sequence generation and word-level embeddings, suitable for longer sentences.

\section{spamGAN}
In this section, we will present the problem set-up, the three components of spamGAN as well as their interactions through a sequential decision making framework. 
\subsection{Problem Set-up}
Let $\mathbb{D_L}$ be the set of reviews labeled spam or non-spam. Given the cost of labeling, we hope to improve classification performance by also using $\mathbb{D_U}$, a significantly larger set of unlabeled reviews\footnote{$\mathbb{D_U}$ includes both spam/non-spam reviews.}. Let $\mathbb{D}=\mathbb{D_L} \cup \mathbb{D_U}$ be a combination of labeled and unlabeled sentences for training\footnote{Training (see Alg.~\ref{alg:spamgan}) can use only $\mathbb{D_L}$ or both $\mathbb{D_L}$ and $\mathbb{D_U}$.}. Each training sentence $y_{1:T}= \{y_1, y_2, \ldots y_t, \ldots, y_T\}$ consists of a sequence of $T$ word tokens, where $y_t \in \mathtt{Y}$ represents the $t^{th}$ token in the sentence and $\mathtt{Y}$ is a corpus of tokens used. For sentences belonging to $\mathbb{D_L}$, we also include a class label belonging to one of the $2$ classes $ \mathfrak{c} \in \mathbb{C}:\{\mathtt{spam}, \mathtt{non\text{-}spam}\}$. 

\begin{figure}
  \includegraphics[width=\linewidth]{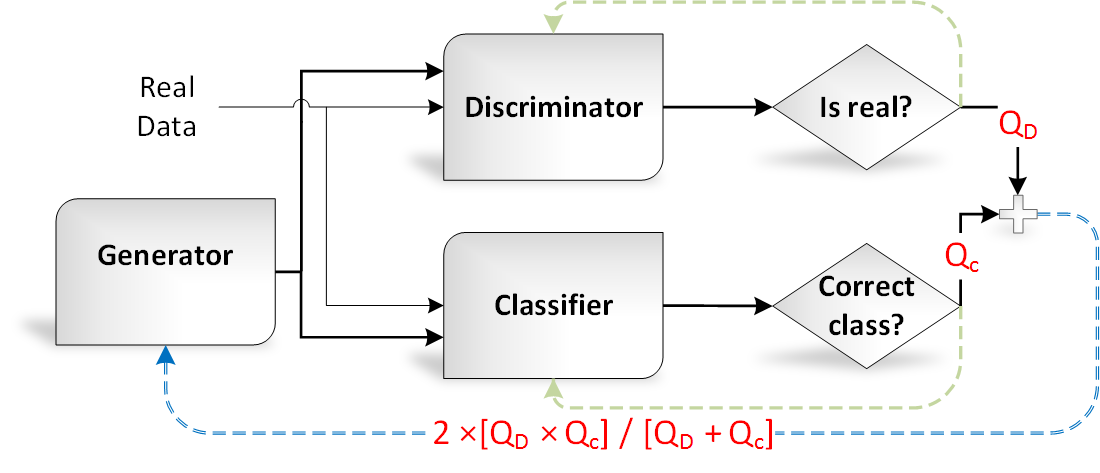}
  \caption{spamGAN Architecture}
  \label{fig:architecture}
\end{figure}

To leverage both the labeled and unlabeled data, we include three components in spamGAN: the generator $\mathcal{G}$, the discriminator $\mathcal{D}$, and the classifier $\mathcal{C}$ as shown in Fig.~\ref{fig:architecture}. The generator, for a given class label, learns to generate new sentences (we call them $\mathtt{fake}$\footnote{Fake sentences are those produced by the generator. Spam sentences are deceptive sentences with class label $\mathtt{spam}$. Generator can generate fake sentences belonging to $\{\mathtt{spam}$ or $\mathtt{non\text{-}spam}\}$ class.} sentences) similar to the real sentences in the train set belonging to the same class. The discriminator learns to differentiate between real and fake sentences, and informs the generator (via rewards) if the generated sentences are unrealistic. This competition between the generator and discriminator improves the quality of the generated sentence. 

We know the class labels for the fake sentences produced by the generator as they are controlled~\cite{hu2017toward}, i.e., constrained by class labels $\{\mathtt{spam}, \mathtt{non\text{-}spam}\}$. The classifier is trained using real labeled sentences from $\mathbb{D_L}$ and fake sentences produced by the generator, thus improving its ability to generalize beyond the small set of labeled sentences. The classifier's performance on fake sentences is also used as feedback to improve the generator: better classification accuracy results in more rewards. While the discriminator and generator are competing, the classifier and generator are mutually bootstrapping. As the $3$ components of spamGAN are trained, the generator produces sentences very similar to the training set while the classifier learns the characteristics of spam and non-spam sentences in order to identify them correctly.

% As shown in the figure the minmax game is played between $G$ and $D$ together~\cite{li2018generative}. The generator is responsible for controlled generation of sentences~\cite{hu2017toward}, i.e., sentences belonging to a particular class. As in SeqGAN~\cite{yu2017seqgan}, we treat the problem of sequence generation as a sequential decision making problem. As shown in Fig. , the generator acts as a reinforcement learning agent, generating the sentence constrained on the class type. The discriminator and classifier evaluate the sentence and give feedback (as rewards) to guide the generator. The classifier [TODO: what is the objective of the classifier?] [TODO: what is the difference in the role between the discriminator and classifier]. [TODO: how is the minmax game played between the 3 components?]

\subsection{Generator}
If $P_R(y_{1:T}, \mathfrak{c})$ is the true joint distribution of sentences $y_{1:T}$ and classes $\mathfrak{c} \in \mathbb{C}$ from the real training set, the generator aims to find a parameterized conditional distribution $\mathcal{G}(y_{1:T} | z, c,\theta_g)$ that best approximates the true distribution. The generated fake sentence is conditioned on the network parameters $\theta_g$, noise vector $z$, and class label $c$, which are sampled from the prior distribution $P_z$ and $P_{\mathfrak{c}}$, respectively. $z$ and $c$ together make up the context vector. The context vector is concatenated to the generated sentence at every timestep~\cite{tuan2018improving}, ensuring that the actual class labels for each generated fake sentence is retained.

 While sampling from $\mathcal{G}(y_{1:T} | z, c,\theta_g)$, the word tokens are generated auto-regressively, decomposing the distribution over token sequences into the ordered conditional sequence,

\vspace{-2mm}
\begin{equation}
\vspace{-2mm}
\mathcal{G}(y_{1:T} | z, c, \theta_g) = \prod_{t=1}^{T} \mathcal{G}(y_t| y_{1:t-1}, z, \mathfrak{c}, \theta_g)
\end{equation}

During pre-training, we use batches of real sentences from $\mathbb{D}$ and minimize the cross-entropy of the next token conditioned on the preceding ones. Specifically, we minimize the loss (Eqn.~\ref{eqn:genmlegrad}) over real sentence-class pairs $(y_{1:T}, \mathfrak{c})$ from $\mathbb{D_L}$ as well as unlabeled real sentences from $\mathbb{D_U}$ with randomly-assigned class labels drawn from the class prior distribution.

\vspace{-2.5mm}
\begin{equation}
\vspace{-2.0mm}
\mathcal{L}^{\mathcal{G}}_{MLE} = - \sum_{t=1}^{T} \log{\mathcal{G}(y_t | y_{1:t-1}, z, \mathfrak{c}, \theta_g)} \label{eqn:genmlegrad}
\end{equation}

  During adversarial training, we treat sequence generation as a sequential decision making problem~\cite{yu2017seqgan}. The generator acts as a reinforcement learning agent and is trained to maximize the expected rewards using policy gradients, where the rewards are feedback obtained from the discriminator and classifier for the generated sentences (See Sec.~\ref{sec:rl}). For implementation, we use a unidirectional multi-layer recurrent neural network with gated recurrent units as the base cell to represent the generator. %However, once could use sophisticated sentence generation architecture such as Transformer with self-attention~\cite{Vaswani2017}.

\subsection{Discriminator}

The discriminator $\mathcal{D}$, with parameters $\theta_d$ predicts if a sentence is real (sampled from $P_R$) or fake (produced by the generator) by computing a probability score $\mathcal{D}(y_{1:T}|\theta_d)$ that the sentence is real. Like \cite{tuan2018improving} instead of computing the score at the end of the sentence, the discriminator produces scores for every timestep $\qthd$, which are then averaged to produce the overall score. 

\vspace{-2.5mm}
\begin{equation}
\vspace{-2.0mm}
\begin{split}
\mathcal{D}(y_{1:T}|\theta_d) & = \frac{1}{T} \sum_{t=1}^T  \qthd \label{eqn:dqvalue}
\end{split}
\end{equation}

$\qthd$ is the intermediate score for timestep $t$ and is based solely on the preceding partial sentence, $y_{1:t}$. In a setup reminiscent of $Q$-learning, we consider $\qthd$ to be the estimated value for the state $s=y_{1:t-1}$ and action $a=y$. %As the notation suggests, the discriminator scores $\qthd$ are shown in \cite{tuan2018improving} to serve as estimators for the true state-action values. 
Thus, the discriminator provides estimates for the true state-action values without the additional computational overhead of using MCTS rollouts.  

We train the discriminator like traditional GANs by maximizing the score $\mathcal{D}(y_{1:T}|\theta_d)$ for real sentences and minimizing it for fake ones. This is achieved by minimizing the loss $\mathcal{L^{(D)}}$,

%\vspace{-1.5mm}
\begin{equation}
%\vspace{-1.0mm}
\begin{split}
\hspace{-2.5mm} \mathcal{L^{(D)}} \hspace{-1.5mm} = \hspace{-2.5mm} & \hspace{-1.5mm} \mathop{\mathbb{E}}_{y_{1:T} \sim P_R} \hspace{-3.5mm}  -\sbr{\log \mathcal{D}(y_{1:T}|\theta_d)} \hspace{-0.5mm} + \hspace{-1.5mm} \mathop{\mathbb{E}}_{y_{1:T} \sim \mathcal{G}}\hspace{-1.5mm} - \hspace{-0.5mm} \sbr{\log{(1 \hspace{-1.5mm}- \hspace{-1.5mm}\mathcal{D}(y_{1:T}|\theta_d))}} \label{eqn:dloss}
\end{split}
\end{equation}

We also include a discrimination critic $\mathcal{D}_{crit}$~\cite{konda2000actor} which is trained to approximate the score $\qthd$ from the discriminator network, for the next token $y_t$ based on the preceding partial sentence $y_{1:t-1}$. The approximated score $\vthd$ will be used to stabilize policy gradient updates for the generator during adversarial training. 

%\vspace{-1mm}
\begin{equation}
%\vspace{-1mm}	
 \vthd = \mathop{\mathbb{E}}_{y_t} \sbr{\qthd}
\end{equation}

$\mathcal{D}_{crit}$ is trained to minimize the sequence mean-squared error between $\vthd$ and the actual score $\qthd$.

%\vspace{-1mm}
\begin{equation}
%\vspace{-1mm}
\begin{split}
\mathcal{L^{(D_\text{crit})}} & = \mathop{\mathbb{E}}_{y_{1:T}} \sum_{t=1}^{T} \norm{\qthd - \vthd}^2 
\label{eqn:dcriticloss}
\end{split}
\end{equation}
% Gradient of critic loss 
% \nabla_{\theta_{\mathcal{D}_{crit}}} \mathcal{L^{(D_\text{crit})}} & =  \mathop{\mathbb{E}}_{y_{1:t-1}}\sbr{ \nabla_{\theta_{dcrit}} \sum_{t=1}^{T}  \norm{{\qthd - V_{D}(y_{1:t-1}|\theta_{dcrit})}^2}}

 The discriminator network is implemented as a unidirectional Recurrent Neural Network (RNN) with one dense output layer which produces the probability that a sentence is real at each timestep, i.e., $\qthd$. For the discrimination critic, we have a additional output dense layer (different from the one that computes $\qthd$) attached to the discriminator RNN, which estimates $V_{\mathcal{D}}(y_{1:t-1})$ for each timestep.

\subsection{Classifier}

Given a sentence $y_{1:T}$, the classifier $\mathcal{C}$ with parameters $\theta_c$ predicts if the sentence belongs to class $c \in \mathbb{C}$. Like the discriminator, it assigns a prediction score at each timestep $\qthc$ for the partial sentence $y_{1:t}$, which identifies the probability the sentence belongs to class $c$. The intermediate scores are then averaged to produce the overall score:
%\vspace{-1.0mm}
\begin{equation}
%\vspace{-1.5mm}
\begin{split}
\mathcal{C}(y_{1:T}, c|\theta_c) & = \frac{1}{T} \sum_{t=1}^T \qthc  \label{eqn:cqvalue}
\end{split}
\end{equation}

The classifier loss $\mathcal{L^{C}}$ is based on: 1) $\mathcal{L^{(C_{\text{R}})}}$, the cross-entropy loss on true labeled sentences computed using the overall classifier sentence score; 2) $\mathcal{L^{(C_{\text{G}})}}$ the loss for the fake sentences. Fake sentences are considered as potentially-noisy training examples, so we not only minimize cross-entropy loss but also include Shannon entropy $\mathcal{H}(\mathcal{C}(c|y_{1:T},\theta_C))$. 

\begin{equation}
\vspace{-2.0mm}
\begin{split}
\mathcal{L^{C}} & = \mathcal{L^{(C_{\text{R}})}} + \mathcal{L^{(C_{\text{G}})}}\\ \label{eqn:closs} 
%\nabla_{\theta_c} \mathcal{L^{C}} & = \nabla_{\theta_c} \mathcal{L^{(C_{\text{R}})}} + \nabla_{\theta_c} \mathcal{L^{(C_{\text{G}})}} 
\end{split}
\vspace{-2.0mm}
\end{equation}
\vspace{-3.0mm}
\begin{equation*}
\begin{split}
\mathcal{L^{(C_{\text{R}})}}& = \mathop{\mathbb{E}}_{(y_{1:T}, c) \sim P_R(y, \mathfrak{c})}\sbr{-\log \mathcal{C}(c|y_{1:T},\theta_c)} \\ 
\mathcal{L^{(C_{\text{G}})}}  & = \mathop{\mathbb{E}}_{ c \sim P_c, y_{1:T} \sim \mathcal{G}} \sbr{-\log \mathcal{C}(c|y_{1:T},\theta_c) -  \beta \mathcal{H}(\mathcal{C}(c|y_{1:T},\theta_C))}
\end{split}
\end{equation*}

%\vspace{-2.5mm}
%\begin{equation*}
%\vspace{-1mm}
%\begin{split}
%\mathcal{L^{(C_{\text{G}})}}  & = \mathop{\mathbb{E}}_{ c \sim P_c, y_{1:T} \sim \mathcal{G}} \sbr{-\log \mathcal{C}(c|y_{1:T},\theta_c) -  \beta \mathcal{H}(\mathcal{C}(c|y_{1:T},\theta_C))} 
%\end{split}
%\end{equation*}

In $\mathcal{L^{(C_{\text{G}})}}$, $\beta$, the balancing parameter, influences the impact of Shannon entropy. Including $\mathcal{H}(\mathcal{C}(c|y_{1:T},\theta_C))$, for minimum entropy regularization~\cite{hu2017toward}, allows the classifier to predict classes for generated fake sentences more confidently. This is crucial in reinforcing the generator to produce sentences of the given class during adversarial training. 

Like in discriminator, we include a classification critic $\mathcal{C}_{crit}$ to estimate the classifier score $\qthc$ for $y_{t}$ based on the preceding partial sentence $y_{1:t-1}$, %given by $\vthc$,

\begin{equation}
\vthc = \mathop{\mathbb{E}}_{y_t}[\qthc]
\end{equation}

 The implementation of the classifier is similar to the discriminator. We use a unidirectional recurrent neural network with a dense output layer producing the predicted probability distribution over classes $\mathfrak{c} \in \mathbb{C}$. The classification critic is also an alternative head off the classifier RNN with an additional dense layer estimating $\vthc$ for each timestep. We train this classifier critic by minimizing $\mathcal{L^{(C\text{crit})}}$,

%\begin{equation}
%\begin{split}
%\nabla_{\theta_c} \mathcal{L^{(c)}}  & = \mathop{\mathbb{E}}_{(y, c) \sim P_R(y, \mathcal{c})} \nabla_{\theta_c} \sbr{-\log \mathcal{C}(c|y_{1:T},\theta_c)}+ \\
%&  \mathop{\mathbb{E}}_{(y, c) \sim \mathcal{G}, P_c} \nabla_{\theta_c} \sbr{-\log \mathcal{C}(c|y_{1:T},\theta_c)} - \\
%&   \nabla_{\theta_c} \beta \mathcal{H}(\mathcal{C}(c|y_{1:T},\theta_c))
%\end{split}
%\end{equation}\label{eqn:clossgradient}

%In order to estimate $V(y_{1:t-1},c)$ in Eqn~\ref{eqn:deltag}, we also use the classifier critic network. 

\vspace{-2.5mm}
\begin{equation}
\begin{split}
\mathcal{L^{(C_\text{crit})}} & = \mathop{\mathbb{E}}_{y_{1:T}} \sum_{t=1}^{T} \norm{\qthc - \vthc}^2  \label{eqn:ccriticloss}
\end{split}
\end{equation}

% Gradient of classifier critic loss
%\nabla_{\theta_{\mathcal{C}_{crit}}} \mathcal{L^{(C_\text{crit})}} & =  \mathop{\mathbb{E}}_{y_{1:t-1}}\sbr{ %\nabla_{\theta_{ccrit}} \sum_{t=1}^{T}  \norm{{\qthc - V_{C}(y_{1:t-1}, c|\theta_{dcrit})}^2}}

\subsection{Reinforcement Learning Component}\label{sec:rl}

We consider a sequential decision making framework, in which the generator acts as as a reinforcement learning agent. The current state of the agent is the generated tokens $s_t = y_{1:t-1}$ so far. The action $y_{t}$ is the next token to be generated, which is selected based on the stochastic policy $\mathcal{G}(y_t| y_{1:t-1}, z, c, \theta_g)$.
The reward the agent receives for the generated sentence $y_{1:T}$ of a given class $\mathfrak{c}$ is determined by the discriminator and classifier. Specifically, we take the overall scores $\mathcal{D}(y_{1:T}| \theta_d)$ (Eqn.\ref{eqn:dqvalue}) and $\mathcal{C}(y_{1:T}, c| \theta_c)$ (Eqn.~\ref{eqn:cqvalue}) and blend them in a manner reminiscent of the F1 score, producing the sentence reward,
\begin{equation}
R(y_{1:T}) = 2\cdot \frac{\mathcal{D}(y_{1:T} | \theta_d) \cdot \mathcal{C}(y_{1:T}, c | \theta_c)}{\mathcal{D}(y_{1:T} | \theta_d) + \mathcal{C}(y_{1:T}, c | \theta_c)} \label{blendedreturn}
\end{equation}

This reward $R(y_{1:T})$ is for the entire sentence delivered during the final timestep, with reward for every other timestep being zero~\cite{tuan2018improving}. Thus, the generator agent seeks to maximize the expected reward, given by, 

\begin{equation}
\begin{split}
\mathcal{L^{(G)}} & = \mathop{\mathbb{E}}_{y_{1:T} {\sim} \mathcal{G}}\sbr{R(y_{1:T})}
\end{split}
\label{eqn:expectreward}
\end{equation}

To maximize $\mathcal{L^{(G)}}$, the generator parameters $\theta_g$ are updated via policy gradients~\cite{sutton2000policy}. Specifically, we use the advantage actor-critic method to solve for optimal policy~\cite{konda2000actor}. The expectation in Eqn.~\ref{eqn:expectreward} can be re-written using rewards for intermediate time-steps from the discriminator and classifier. The intermediate scores from the discriminator, $\qthd$ and the classifier, $\qthc$, are combined as shown in Eqn.~\ref{eqn:rewardspertimestep} 
and the combined values serve as estimators for $Q(y_{1:t}, c)$, the expected reward for sentence $y_{1:t}$. To reduce variance in the gradient estimates, we replace $Q(y_{1:t}, c)$ by the advantage function $Q(y_{1:t}, c) - V(y_{1:t-1}, c)$, where $V(y_{1:t-1}, c)$ is given by Eqn.~\ref{eqn:rewardspertimestep}. We use $\alpha = T - t$ in Eqn.~\ref{eqn:deltag} to increase the importance of initially-generated tokens while updating $\theta_g$. $\alpha$ is a linearly-decreasing factor which corrects the relative lack of confidence in the initial intermediate scores from the discriminator and classifier.
% Similarly, blending the state values $V_\mathcal{D}(y_{1:t})$ and $V_{\mathcal{C}} (y_{1:t}, \mathfrak{c})$ provides an estimate of the true value function. 

\begin{algorithm}[t]
\small{
\caption{spamGAN}
\label{alg:spamgan}
\textbf{Input}: Labeled dataset $\mathbb{D_L}$, Unlabeled dataset $\mathbb{D_U}$ \\
\textbf{Parameters}: Network parameters $\theta_g\; \theta_d \; \theta_c \; \theta_{dcrit} \; \theta_{ccrit}$ \\
%\textbf{Output}: Your algorithm's output
 %[1] enables line numbers
Perform pre-training as described in Sec.~\ref{sec:pre}\\
\For{$\mathtt{Training\text{-}epochs}$}{
\For{$\mathtt{G\text{-}Adv\text{-}epochs}$}{
sample batch of classes $\mathfrak{c}$ from $\sim P(c)$ \\
generate batch of fake sequences $y_{1:T} \sim \mathcal{G}$ given $\mathfrak{c}$\\
\For{$t \in 1:T$}{
compute $Q(y_{1:t}, c)$, $V(y_{1:t-1},c)$ using Eqn.~\ref{eqn:rewardspertimestep}
}
update $\theta_g$ using policy gradient $\nabla_{\theta_g} \mathcal{L^{(G)}}$ in Eqn.~\ref{eqn:deltag}\\
}
\For{$\mathtt{G\text{-}MLE\text{-}epochs}$}{
sample batch of real sentences from $\mathbb{D_L}$, $\mathbb{D_U}$ \\
Update $\theta_g$ using MLE in Eqn.~\ref{eqn:genmlegrad} \\
}

\For{$\mathtt{D\text{-}epochs}$}{
sample batch of real sentences from $\mathbb{D_L}$, $\mathbb{D_U}$ \\
sample batch of fake sentences from $\mathcal{G}$  \\
update discriminator using $\nabla_{\theta_d} \mathcal{L^{(D)}}$ from Eqn.~\ref{eqn:dloss}\\
compute $\qthd, \vthd$ for fake sentcs \\
update $\mathcal{D}_{\text{crit}}$ using $\nabla_{\theta_{dcrit}} \mathcal{L^{(D\text{crit})}}$ from Eqn.~\ref{eqn:dcriticloss} \\
}
\For{$\mathtt{C\text{-}epochs}$}{
sample batch of real sentences-class pairs from $\mathbb{D_L}$ \\
sample batch of fake sentence-class pairs from $\mathcal{G}$ \\
update classifier using $\nabla_{\theta_c} \mathcal{L^{(C)}}$ from Eqn.~\ref{eqn:closs} \\
\hspace{-2mm} compute \hspace{-1mm} $\qthc,\hspace{-0.5mm}\vthc$ on fake sents \\
update $\mathcal{C}_{\text{crit}}$ using $\nabla_{\theta_{ccrit}} \mathcal{L^{(C\text{crit})}}$ from Eqn.~\ref{eqn:ccriticloss}
}
}
}
\vspace{-1.5mm}
\end{algorithm}

\begin{equation}
\begin{split}
& Q(y_{1:t}, c)  = 2 \cdot \frac{\qthd \cdot \qthc}
{\qthd +\qthc}\\
& V(y_{1:t-1},c) = 2 \cdot \frac{\vthd \cdot  \vthc}
{\vthd + \vthc} 
\label{eqn:rewardspertimestep} 
\end{split}
\end{equation}

During adversarial training, we perform gradient ascent to update the generator using the gradient equation shown below, 

\vspace{-2.5mm}
\begin{align}
\nabla_{\theta_g} \mathcal{L^{(G)}}  = \mathop{\mathbb{E}}_{y_{1:T}} \sum_t^T & \alpha \sbr{Q(y_{1:t}, c) - V(y_{1:t-1}, c)} \nonumber \\
\vspace{-2.5mm}
  & \times \nabla_{\theta_g} \log \mathcal{G}(y_t | y_{1:t-1}, z, c, \theta_g) 
  \label{eqn:deltag}
  \vspace{-3.5mm}
\end{align}
\vspace{-5.5mm}

%\begin{equation*}
%\vspace{-2.5mm}
%\begin{split}
%V(y_{1:t-1},c)& = 2 \cdot \frac{V_{\mathcal{D}}(y_{1:t-1}) \cdot  V_{\mathcal{C}}(y_{1:t-1},c)}
%{V_{\mathcal{D}}(y_{1:t-1}) + V_{\mathcal{C}}(y_{1:t-1},c)} 
%\end{split}
%\end{equation*}

\begin{table*}[t]
\small{
	\caption{Accuracy (Mean $\pm$ Std) for Different \% Labeled Data}
	\vspace{-2.5mm}
	\begin{center}
\begin{tabular}{l*{6}{l}}
	\toprule
			%& \multicolumn{6}{c}{Training \%} \\
	Method  &10\% Labeled &30\%&50\%&70\%&90\%&100\%\\
	\midrule 
	spamGAN-0\%   & 0.700 $\pm$ 0.02 & 0.811 $\pm$ 0.02 & 0.838 $\pm$ 0.01 &	0.845 $\pm$ 0.01 &	0.852 $\pm$ 0.02 &	0.862 $\pm$ 0.01  \\
	spamGAN-50\%  & 0.678 $\pm$ 0.03 &	0.797 $\pm$ 0.03 &	0.839 $\pm$ 0.02 &	0.845 $\pm$ 0.02 &	0.857 $\pm$ 0.02 &	0.856 $\pm$ 0.01 \\
	spamGAN-70\%  & 0.695 $\pm$ 0.05 &	0.780 $\pm$ 0.03 &	0.828 $\pm$ 0.02 &	0.850 $\pm$ 0.01 &	0.841 $\pm$ 0.02 &	0.844 $\pm$ 0.02 \\
	spamGAN-100\%  & 0.681 $\pm$ 0.02 &	0.783 $\pm$ 0.02 &	0.831 $\pm$ 0.01	&0.837 $\pm$ 0.01	& 0.843 $\pm$ 0.02 &	0.845 $\pm$ 0.01\\
	Base classifier  & 0.722 $\pm$ 0.03 &	0.786 $\pm$ 0.02 &	0.791 $\pm$ 0.02 &	0.829 $\pm$ 0.01 &	0.824 $\pm$ 0.02 &	0.827 $\pm$ 0.02\\
	DRI-RCNN & 0.647 $\pm$ 0.10 & 0.757 $\pm$ 0.01 & 0.796 $\pm$ 0.01 & 0.834 $\pm$ 0.18 & 0.835 $\pm$ 0.02 & 0.846 $\pm$ 0.01 \\
	RCNN  & 0.538 $\pm$ 0.09 & 0.665 $\pm$ 0.14 & 0.733 $\pm$ 0.09 & 0.811 $\pm$ 0.03 & 0.834 $\pm$ 0.02 & 0.825 $\pm$ 0.02 \\
	Co-Train (Naive Bayes) & 0.655 $\pm$ 0.01  & 0.740 $\pm$ 0.01 & 0.738 $\pm$ 0.02 & 0.743 $\pm$ 0.01 & 0.754 $\pm$ 0.01 & 0.774 $\pm$ 0.01 \\
	PU Learn (Naive Bayes) & 0.508 $\pm$ 0.02 & 0.713 $\pm$ 0.03 & 0.816 $\pm$ 0.01 & 0.826 $\pm$ 0.01 & 0.838 $\pm$ 0.02 & 0.843 $\pm$ 0.02 \\
	\bottomrule
\end{tabular}
\label{tab:accuracy}
\end{center}
}
\vspace{-2.5mm}
\end{table*}

\subsection{Pre-Training}\label{sec:pre}

Before beginning adversarial training, we pre-train the different components of spamGAN. The generator $\mathcal{G}$ is pre-trained using maximum likelihood estimation (MLE)~\cite{grover2018flow} by updating the parameters via Eqn~\ref{eqn:genmlegrad}. Once the generator is pre-trained, we take batches of real sentences from the labeled dataset $\mathbb{D_L}$, the unlabeled dataset $\mathbb{D_U}$ and fake sentences sampled from $\mathcal{G}(y_{1:T} | z, c, \theta_g)$ to pre-train the discriminator minimizing the loss $\mathcal{L^{(D)}}$ in Eqn.~\ref{eqn:dloss}. The classifier $\mathcal{C}$ is pre-trained solely on real sentences from the labeled dataset $\mathbb{D_L}$. It is trained to minimize the cross-entropy loss $\mathcal{L^{(C_{\text{R}})}}$ on real sentences and their labels. The critic networks $\mathcal{D_{\text{crit}}}$ and $\mathcal{C_{\text{crit}}}$ are trained by minimizing their loses $\mathcal{L^{(D\text{crit})}}$ (Eqn.~\ref{eqn:dcriticloss}) and $\mathcal{L^{(C\text{crit})}}$ (Eqn.~\ref{eqn:ccriticloss}). Such pre-training addresses the problem of mode collapse~\cite{guo2018long} to a satisfactory extent.

%\begin{algorithm}[b]
%\caption{Pre-Training}
%\label{alg:pre-training}
%\begin{algorithmic}[1] %[1] enables line numbers
%\STATE Initialize Generator $\mathcal{G}$, Discriminator $\mathcal{D}$, Discriminatation Critic $\mathcal{D_{\text{crit}}}$, Classifier $\mathcal{C}$, Classification Critic $\mathcal{C_{\text{crit}}}$
%\STATE Pretrain $\mathcal{G}$:\: minimize MLE using real sentences from $\mathbb{D}$ 
%\STATE Pretrain $\mathcal{D}$: \: minimize $\mathcal{L^{(D)}}$ using real sentences from $\mathbb{D}$, fake sentences from $\mathcal{G}$
%\STATE Pretrain $\mathcal{D_{\text{crit}}}$: \: minimize $\mathcal{L^{(D\text{crit})}}$ using output of the discriminator for fake sentences from $\mathcal{G}$
%\STATE Pretrain $\mathcal{C}$:  \: minimize cross-entropy loss using real labeled sentences from $\mathbb{D}_l$ 
%\STATE Pretrain $\mathcal{C_{\text{crit}}}$: \:  minimize $\mathcal{L^{(C\text{crit})}}$ using output of the classifier for fake sentences from $\mathcal{G}$
%\end{algorithmic}
%\end{algorithm}

\subsection{spamGAN algorithm}
 Alg.~\ref{alg:spamgan} describes spamGAN in detail. After pre-training, we perform adversarial training for $\mathtt{Training\text{-}epochs}$ (Lines $4$-$25$). We create a batch of fake sentences using generator $\mathcal{G}$ by sampling classes $c$ from prior $P_c$ (Lines $6$-$7$). We compute $Q(y_{1:t}, c)$, $V(y_{1:t-1},c)$ using Eqn.~\ref{eqn:rewardspertimestep} for every timestep (Line $9$). The generator is then updated using policy gradient in Eqn.~\ref{eqn:deltag} (Line $10$). This process is repeated for $\mathtt{G\text{-}Adv\text{-}epochs}$. Like \cite{Li2017} the training robustness is greatly improved when the generator is updated using MLE via Eqn~\ref{eqn:genmlegrad} on sentences from $\mathbb{D}$ (Lines $11$-$13$). We then train the discriminator using real sentences from $\mathbb{D_L}$, $\mathbb{D_U}$ as well as fake sentences from the generator (Lines $15$-$16$). The discriminator is updated using Eqn.~\ref{eqn:dloss} (Line $17$). We also train the discrimination critic, by computing $\qthd, \vthd$ for the fake sentences and updating the gradients using Eqn.~\ref{eqn:dcriticloss} (Line $18$-$19$). This process is repeated for $\mathtt{D\text{-}epochs}$. We perform a similar set of operations for the classifier (Lines $20$-$25$).

\begin{table*}[t]
\small{
	\caption{F1-Score (Mean $\pm$ Std) for Different \% Labeled Data}
	\vspace{-2.5mm}
	\begin{center}
\begin{tabular}{l*{6}{l}}
	\toprule
			%& \multicolumn{6}{c}{Training \%} \\
	Method &10\% Labeled &30\%&50\%&70\%&90\%&100\%\\
	\midrule 
	spamGAN-0\%   & 0.718 $\pm$ 0.02	& 0.812 $\pm$ 0.02 &	0.840 $\pm$ 0.01 &	0.848 $\pm$ 0.02 &	 0.854 $\pm$ 0.02 &	0.868 $\pm$ 0.01  \\
	spamGAN-50\%  & 0.674 $\pm$ 0.05	&0.797 $\pm$ 0.03 &0.843 $\pm$ 0.01	 &0.848 $\pm$ 0.02	& 0.860 $\pm$ 0.02 &	0.863 $\pm$ 0.01 \\
	spamGAN-70\%  & 0.702 $\pm$ 0.05	&0.784 $\pm$ 0.03	&0.830 $\pm$ 0.02	&0.856 $\pm$ 0.01	&0.848 $\pm$ 0.02 &	0.854 $\pm$ 0.01 \\
	spamGAN-100\%  & 0.684 $\pm$ 0.03 &	0.788 $\pm$ 0.03 &	0.839 $\pm$ 0.02 &	0.844 $\pm$ 0.01	&0.846 $\pm$ 0.02 &	0.850 $\pm$ 0.01 \\
	Base classifier  & 0.731 $\pm$ 0.03	&0.795 $\pm$ 0.03	&0.803 $\pm$ 0.02	&0.829 $\pm$ 0.01	&0.832 $\pm$ 0.02	&0.838 $\pm$ 0.02 \\
	DRI-RCNN  & 0.632 $\pm$ 0.07 & 0.754 $\pm$ 0.02 & 0.779 $\pm$ 0.00 & 0.812 $\pm$ 0.03 & 0.817 $\pm$ 0.03 & 0.833 $\pm$ 0.02\\
	RCNN   & 0.638 $\pm$ 0.01 & 0.715 $\pm$ 0.01 & 0.754 $\pm$ 0.02 & 0.776 $\pm$ 0.05 & 0.820 $\pm$ 0.03 & 0.833 $\pm$ 0.02\\
	Co-Train (Naive Bayes) & 0.637 $\pm$ 0.02 & 0.698 $\pm$ 0.01 &  0.680 $\pm$ 0.02 & 0.677 $\pm$ 0.01& 0.712 $\pm$ 0.01 & 0.726 $\pm$ 0.01 \\
	PU-Learn (Naive Bayes) & 0.050 $\pm$ 0.02 & 0.636 $\pm$ 0.05 & 0.815 $\pm$ 0.02 & 0.837 $\pm$ 0.02 & 0.844 $\pm$ 0.02 & 0.852 $\pm$ 0.01 \\
	\bottomrule
\end{tabular}
\end{center}
\label{tab:f1}
}
\vspace{-1.5mm}
\end{table*}

\section{Experiments}

%Due to the increased importance of on online customer reviews to purchase decision-making, individuals and businesses are incentivized to game the system by constructing reviews with the intent to deceive other potential customers. This is known as deceptive opinion spam. Human judges are ineffective at separating deceptive reviews from true ones (achieving only 57\% accuracy in Ott, 2011), and so algorithmic solutions would be impactful.

%As deceptive opinions are by their very nature created covertly, 
We use the TripAdvisor labeled dataset~\cite{Ott2011FindingDO}~\footnote{http://myleott.com/op-spam.html}, consisting of 800 truthful reviews on Chicago hotels as well as $800$ deceptive reviews obtained from Amazon Mechanical Turk. We remove a small number of duplicate truthful reviews, to get a balanced labeled dataset of 1596 reviews. We augment the labeled set with $32,297$ unlabeled TripAdvisor reviews for Chicago hotels~\footnote{http://times.cs.uiuc.edu/~wang296/Data/index.html}. All reviews are converted to lower-case and tokenized at word level, with a vocabulary $\mathtt{Y}$ of $10000$. The maximum sequence length is $T=128$ words, close to the median review length of the full dataset.
 
$\mathtt{Y}$ also includes tokens $\mathtt{\braket{start}}$, $\mathtt{\braket{end}}$, $\mathtt{\braket{unk}}$, and $\mathtt{\braket{pad}}$.  $\mathtt{\braket{start}}$, $\mathtt{\braket{end}}$ are added to the beginning, end of each sentence. Sentences smaller than $T$ are padded with $\mathtt{\braket{pad}}$ while longer ones are truncated, ensuring a consistent sentence length. $\mathtt{\braket{unk}}$ replaces out-of-vocabulary words.

In spamGAN, the generator consists of 2 GRU layers of 1024 units each and an output dense layer providing logits for the $10,000$ tokens. The generator, discriminator and classifier are trained using ADAM optimizer. All use variational dropout=$0.5$ between recurrent layers and word embeddings with dimension $50$. For generator, learning rate = $0.001$, weight decay =$1\times10^{-7}$. Gradient clipping is set to a maximum global norm of $5$. 
The discriminator contains 2 GRU layers of 512 units each and a dense layer with a single scalar output and sigmoid activation. The discrimination critic is implemented as an alternative dense layer. Learning rate =$0.0001$ and weight decay =$1 \times 10^{-4}$. The classifier is similar to discriminator. We set balancing coefficient $\beta=1$. The train time of spamGAN using a Tesla P4 GPU was $\sim 1.5$ hrs.

\begin{figure}
\vspace{-1.5mm}
  \includegraphics[width=\linewidth]{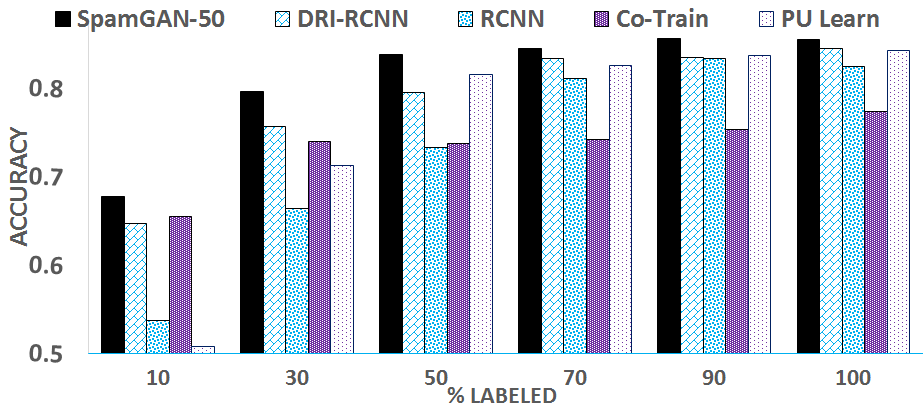}
  \caption{Comparison of spamGAN-50 with Other Approaches}
  \label{fig:accuracy}
  \vspace{-1.5mm}
\end{figure}

%\subsection{Experimental Setup}

We use a $80-20$ train-test split on labeled data. 
We compare spamGAN with $2$ supervised methods which use recurrent networks: 1) DRI-RCNN~\cite{zhang2018dri}; 2) RCNN~\cite{lai2015recurrent} as well as $2$ semi-supervised methods: 3) Co-Training~\cite{li2011learning} with Naive Bayes classifier; 4) PU Learning~\cite{hernandez2013using} with Naive Bayes (SVM performed poorly) using only spam and unlabeled reviews.

We conduct experiments with $10, 30, 50, 70, 90, 100\%$ of labeled data. To analyze the impact of unlabeled data, we show different versions: spamGAN-0 (no unlabeled data), spamGAN-50 (50\% unlabeled data), spamGAN-70 (70\% unlabeled) and spamGAN-100. Co-Train, PU-Learn results are for $50\%$ unlabeled data. We also show the performance of our base classifier (without generator, discriminator, trained on real labeled data to minimize $\mathcal{L^{(C_{\text{R}})}}$). All experiments are repeated $10$ times and the mean, standard deviation are reported. %For selection of hyperparameters as well as of the best-performing model during adversarial training in spamGAN, we evaluate classification metrics on the subset of the labeled training set which was not included in the training subset.

\subsubsection{Influence of Labeled Data}

Table.~\ref{tab:accuracy} shows the classification accuracy of the different models on the test set. SpamGAN models, in general, outperform other approaches, especially when the \% of labeled data is limited. When we merely use $10\%$ of labeled data, spamGAN-0, spamGAN-50, spamGAN-70, spamGAN-100 achieve an accuracy of $0.70, 0.678, 0.695, 0.681$, respectively, which is higher than supervised approaches DRI-RCNN ($0.647$) and R-CNN ($0.538$) as well as semi-supervised approaches Co-train ($0.655$) and PU-learning ($0.508$). Even without any unlabeled data spamGAN-0 gets good results because the mutual bootstrapping between generator and classifier allows the classifier to explore beyond the small labeled training set using the fake sentences produced by the generator. The accuracy of our base classifier is $0.722$, higher than spamGAN models as GANs needs more samples to train, in general. 

The accuracy of all approaches increases with \% of labeled data. We select spamGAN-50 as a representative for comparison in Fig.~\ref{fig:accuracy}. Though the difference in accuracy between spamGAN-50 and others reduces as the \% of labeled data increases, spamGAN-50 still performs better than others with an accuracy of $0.856$ when all labeled data are considered.

Table.~\ref{tab:f1} shows the F1-score. We can again see that spamGAN-0, spamGAN-50 and spamGAN-70 perform better than the others, especially when the \% of labeled data is small.

\subsubsection{Influence of Unlabeled Data}
\begin{figure}
  \includegraphics[width=\linewidth]{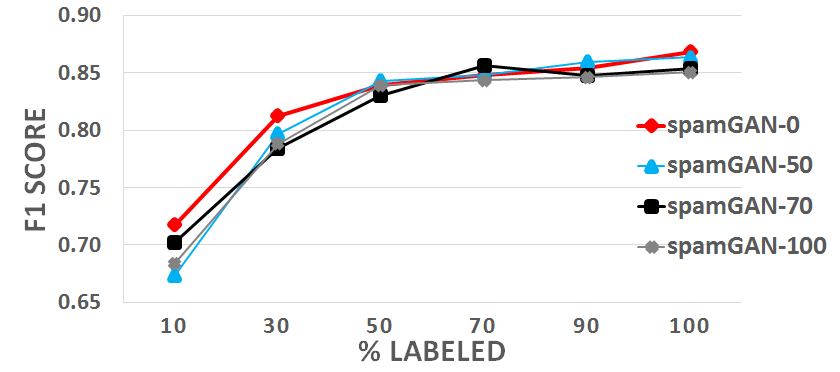}
  \caption{Influence of Unlabeled Data on F1-Score}
  \label{fig:unlabeled}
  \vspace{-1.5mm}
\end{figure}

While unlabeled data is used to augment the classifier's performance, Fig.~\ref{fig:unlabeled} shows that F1-score slightly decreases when the \% unlabeled data increases, especially for spamGAN-100. In our case, as unlabeled data is much larger than the labeled, the generator does not entirely learn the importance of the sentence classes during pre-training (when the unlabeled sentence classes are randomly assigned), which causes problems for the classifier during adversarial training. However, when no unlabeled data is used, the generator easily learns to generate sentences conditioned on classes paving way for mutual bootstrapping between classifier and generator. We can also attribute the drop in performance to the difference in distribution of data between the unlabeled TripAdvisor reviews and the handcrafted reviews from Amazon MechanicalTurk. % where they talk about how adding unlabeled data with a different distribution of classes can impair classification. It might be possible that the different mix between deceptive and truthful reviews in the unlabeled data is sufficiently diverse from the artificial dataset, both in terms of % deceptive / % truthful and in terms of the difference between AmazonTurk-created deceptive reviews vs deceptive reviews “in the wild” is causing this. 

\subsubsection{Perplexity of Generated Sentence}
\begin{figure}
  \includegraphics[width=\linewidth]{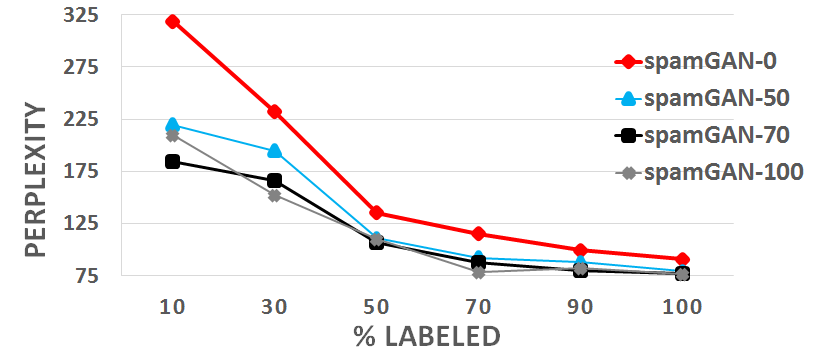}
  \caption{Influence of Unlabeled Data on Perplexity}
  \label{fig:perplexity}
  \vspace{-2mm}
\end{figure}

We also compute the perplexity of the sentences produced by the generator (the lower the value the better). Fig.~\ref{fig:perplexity} shows that as the \% of unlabeled data increases (spamGAN-0 to spamGAN-100), the perplexity of the sentences decreases. SpamGAN-100, SpamGAN-70 achieve a perplexity of $76.4, 76.5$, respectively. Fig.~\ref{fig:unlabeled}, Fig.~\ref{fig:perplexity} show that using unlabeled data improves the generator in producing realistic sentences but does not fully help to differentiate between the classes which again, can be attributed to the difference in the data distribution between the labeled and unlabeled data.

Following is a sample (partial) spam sentence produced by the generator: "Loved this hotel but i decided to the hotel in a establishment didnt look bad ...the palmer house was anyplace that others said in the reviews..". We notice that spam sentences use more conservative choice of words, focusing on adjectives, reviewer, and attributes of the hotel, while non-spam sentences speak more about the trip in general. 

 \vspace{-1.5mm}

\section{Conclusion and Future Work}
We have proposed spamGAN, an approach for detecting opinion spam with limited labeled data. spamGAN, apart from detecting spam, helps to generate reviews similar to the training set. Experiments show that spamGAN outperforms state-of-the-art supervised and semi-supervised techniques when labeled data is limited. While we use TripAdvisor dataset, we plan to conduct experiments on YelpZip data (overcoming the data distribution issue of MechanicalTurk reviews). As the overall spamGAN architecture is agnostic to the implementation details of the classifier, we plan to use a more sophisticated design for classifier than a simple recurrent network.

\bibliographystyle{named}
\bibliography{ijcai19}

\end{document}